\documentclass[runningheads]{llncs}

\usepackage{graphicx, multirow}
\usepackage{mathptmx}
\usepackage{amsmath}
\usepackage[utf8]{inputenc} 
\usepackage[T1]{fontenc}    
\usepackage{hyperref}       
\usepackage{url}            
\usepackage{booktabs}       
\usepackage{nicefrac}       
\usepackage{microtype}      
\usepackage{wrapfig}

\begin{document}
\title{GP-HD: Using Genetic Programming to Generate Dynamical Systems Models for Health Care}

\author{
Mark Hoogendoorn\inst{1} \and Ward van Breda\inst{1} \and Jeroen Ruwaard\inst{2}}


\institute{
$^{1}$Vrije Universiteit Amsterdam, Department of Computer Science\\
De Boelelaan 1081, 1081 HV Amsterdam, the Netherlands\\
\email{\{m.hoogendoorn, w.r.j.van.breda\}@vu.nl}\\
$^{2}$Vrije Universiteit Amsterdam, Department of Clinical, Neuro and Developmental Psychology\\
Van der Boechorstraat 7, 1081 BT Amsterdam, the Netherlands\\
\email{j.j.ruwaard@vu.nl}}

\maketitle

\begin{abstract}
The huge wealth of data in the health domain can be exploited to create models that predict development of health states over time. Temporal learning algorithms are well suited to learn relationships between health states and make predictions about their future developments. However, these algorithms: (1) either focus on learning one generic model for all patients, providing general insights but often with limited predictive performance, or (2) learn individualized models from which it is hard to derive generic concepts. In this paper, we present a middle ground, namely parameterized dynamical systems models that are generated from data using a Genetic Programming (GP) framework. A fitness function suitable for the health domain is exploited. An evaluation of the approach in the mental health domain shows that performance of the model generated by the GP is on par with a dynamical systems model developed based on domain knowledge, significantly outperforms a generic Long Term Short Term Memory (LSTM) model and in some cases also outperforms an individualized LSTM model.
\end{abstract}

\section{Introduction}

Within the domain of health we are faced with an ever increasing amount of data that can be exploited for the benefit of the patient. There are many examples of insights that can be obtained from such data. One case is gaining understanding into how health states evolve over time and how they influence each other. Take the domain of mental health for instance, we might be interested to know what the mood of a depressed patient will be like in a few days time, and how the sleep quality influences this future mood. To derive such patterns from the data, we can apply temporal learning algorithms. When doing so, we have to make a choice in whether we are aiming for a one-size-fits-all model or models per individual (cf.~\cite{hoogendoorn2017machine}). One-size-fits-all models provide generic insights, but often suffer from limited predictive value due to the inherent heterogeneity among patients. Individual models are tailored towards a person, but deriving useful information across all patients is difficult. In addition, limited data is typically available per patient, making it difficult to generate models that generalize well.

A technique that provides a middle ground between generic and individual models is dynamical systems modeling. These models represent the health states of a patient as a numerical value, and specify influence relationships between the states over time by means of difference equations. The equations include parameters that express the strength of the relationships. Hence, equations model the generic relationships and parameter values allow for individual tailoring. Unfortunately until now these kind of models need to be specified by exploiting domain knowledge rather than finding relationships in the data automatically (see e.g.~\cite{abro2016validation,treur2016network,bosse2009generic,bosse2007agent}). This makes the development time consuming, open to interpretation (as theories are often not precise enough to specify a difference equation), and it does not allow one to find new relationships in the data. 

In this paper, we propose an approach that is able to generate dynamical systems models for health using Genetic Programming (GP). While several approaches have been proposed to generate these types of models using GPs (see e.g.\cite{schmidt2009distilling,cao2000evolutionary}) the health domain poses very different challenges. Rather than fitting these models towards a single dataset, in the health domain models should predict well across sets of patients, and should be able to cope with the variability of the patients by means of their parameter values. This has implications for the fitness function used by the GP. To develop such an approach for the health domain, we take an existing GP approach as a basis (cf.~\cite{cao2000evolutionary}). We extend the approach with a fitness function which is based on an evaluation framework for more knowledge driven dynamical systems models  for the health domain (cf.~\cite{van2017assessment}). Overall, this results in an approach to develop accurate and insightful predictive models for the domain of health. We refer to the approach as GP-HD (for GP Health-state Dynamics). We aim to answer the following research question in this paper:
\newline

\noindent \emph{Is the predictive performance of models generated by GP-HD better than state-of-the-art data-driven and knowledge-driven approaches for a real-life case study in the health domain?}
\newline

\noindent To answer this question we evaluate the approach by means of a case study in mental health using a real dataset: forecasting the mood and perceived sleep quality of depressed patients up to three days ahead. We compare the resulting model with an existing knowledge-driven dynamical systems models~\cite{abro2016validation}, individualized Long-Short Term Memory (LSTM) neural networks (cf.~\cite{hochreiter1997long}) and a single generic LSTM model. We also evaluate the influence of several of the key hyperparameters of the system, while fixing others to values reported in prior literature.

This paper is organized as follows. First, we introduce our approach to generate the dynamical systems models. We then present the dataset we use to evaluate our approach in Section 3, while Section 4 provides the experimental setup. The results are presented in Section 5, and finally Section 6 concludes the paper with a discussion.

\section{Approach}

\begin{figure*}[!htb]
\centering
\includegraphics[width=0.8\textwidth]{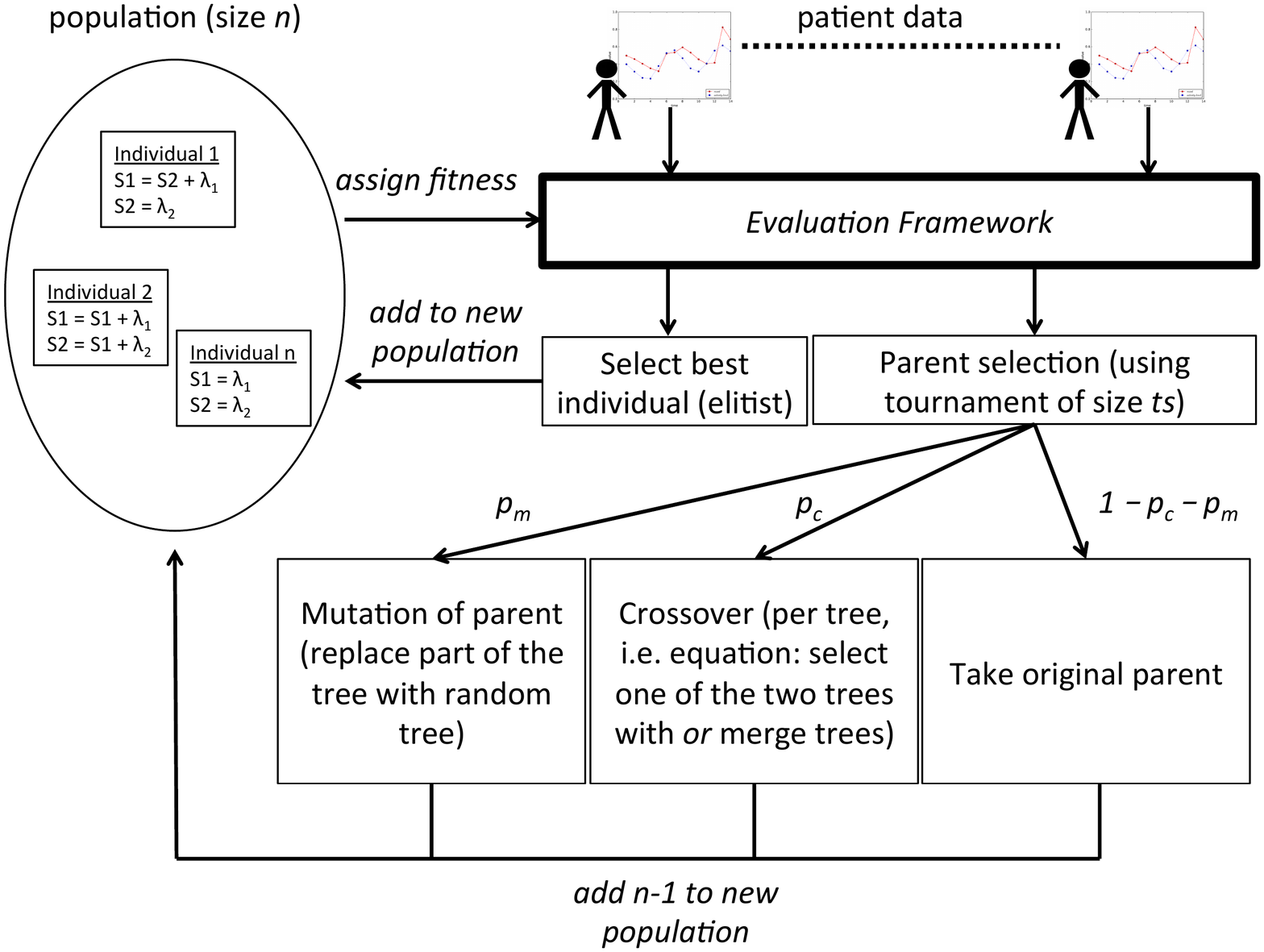}
\caption{Genetic Programming loop for our framework.}
\label{fig:gp_loop} 
\end{figure*}

Before going into detail on GP-HD itself, let us first consider the dynamical systems models we aim to learn. As said, we focus on the development of health states over time, i.e. we are faced with time series per patient and want to predict future values based on historical values. We assume that the equations of the dynamical systems model to make these predictions express the value of a state at time point $t+\Delta t$ based on values of states at $t$, for example: $s_1(t+\Delta t) = s_1(t) + \gamma_1 \cdot s_2(t)$.
The right hand side can contain a combination of states and parameters. As we are interested in not just modeling a single state, but a whole range of states we have a system of such difference equations. Next, we will explain the GP part of the approach, followed by an explanation of the fitness function used.

\subsection{Genetic Programming approach}

Let us first focus on the generic GP loop, shown in Figure~\ref{fig:gp_loop}. This approach is taken from \cite{cao2000evolutionary} with a few minor modifications. \\

\noindent {\bf Individual representation:} We encode the individuals that represent the systems of equations by means of a vector of trees. Each tree $T_i$ in the vector represents the difference equation to compute the new values for state $i$. Assuming $m$ states we have a vector of $m$ trees: $<T_1,\dots,T_m>$. A tree can have mathematical operators as nodes (we use $*$, $+$, and $-$) while the leaves of the tree can either contain a parameter or a state. We put a limit to the depth of the tree ($d_{max}$) and the number of parameters used in the total vector of trees ($\lambda_{max}$).\\

\noindent {\bf Population initialization:} We create an initial population of $n$ individuals. For each tree, we start with an empty tree and randomly select one of the mathematical operators (with probability $p_{op}$) or randomly select one of the terminal with probability $1-p_{op}$ (terminals are states or parameters) in case we have not reached $d_{max}$ yet. In the event that we have reached $d_{max}$ we always select a terminal.\\

\noindent {\bf Parent selection:} We select parent(s) based on a standard tournament selection approach of size $t_s$. \\

\noindent {\bf Variation operators:} According to the standard GP loop (cf.~\cite{eiben2015introduction}) we perform either mutation, crossover, or we copy the original solution. We start by selecting one parent. We copy the parent with a fixed probability $p_r$. The mutation rate depends on the fitness of the parent that has been selected. Assuming the fitness of the parent being $f$ and the fittest one in the population being $f_{max}$ we take the mutation probability as (based on~\cite{cao2000evolutionary}): $p_m = 0.1 + 0.2 \cdot \Big( 1 - \Big( \frac{f}{f_{max}}\Big) \Big)$. Crossover is performed with the remaining probability (i.e. $p_{c}=1-p_m-p_r$).

In case of mutation, a random tree of the parent is selected from which we pick a random node or leaf, and replace it by a randomly generated subtree with a depth such that $d_{max}$ is not exceeded. In case the crossover operator is selected, we select an additional parent using the same parent selection approach. After that, we apply either tree level crossover or vector level crossover with equal probability. In tree level crossover we merge the two parent trees for each position in the vector while in the vector level crossover for each position we randomly select a tree of the two parents with equal probability. Merging of trees is done by selecting a crossover point in each one of the two trees, and substituting the subtree from the crossover point of the first parent by the subtree at the crossover point in the second parent. Since it can result in trees that exceed the maximum depth, we try this $\phi$ times and in case we do not succeed the crossover fails and we select different parents.\\

\noindent {\bf Next Generation:} We create a new population, and add the best individual of the previous generation (elitist approach). We generate children using the variation operators and fill up the new population until we reach the desired population size ($n$).  

\subsection{Determining the Fitness Value}

Determining a fitness value is certainly not trivial for the type of health data we assume. We use the evaluation framework presented in~\cite{van2017assessment} to determine the fitness value of a dynamical systems model. To make the paper self-containing the most important aspects of the approach are presented below.


In the evaluation framework, it is assumed that we have data which expresses the discretized measured values of all relevant states $m$ of $p$ patients over time. Here, the step size equals the step size used in the models ($\Delta t$), i.e. $\{x_i(t_{start}),x_i(t_{start}+\Delta t),\dots,x_i(t_{end})\}$. We split this data up into a training, validation, and test set, taking the first fraction as training data, the middle part as validation set, and the last part as test data. The goal of our models is to predict $n$ time steps ahead and minimize the difference (in terms of the mean squared error) between the values of the states in the model and the real data. Depending on the goal of the model either all measured health states should be predicted well, or a subset thereof (this is a choice for the user of the model in the end). Hence, it is very likely that we want to optimize the predictive performance of multiple health states. This is known as a multi-objective optimization problem. Given the individual candidate model we have (as part of the population in the GP) and the criteria we want to optimize, we need to find values for the set of parameter that are present in the model $\{\gamma_1,\dots,\gamma_k\}$ (where $k \leq \lambda_{max}$). We do this \emph{per patient}. A model with instantiated parameter values is called a \emph{model instance}. Since this is a multi-objective optimization problem, multiple model instances can result that reside on the Pareto front. Each of these model instances obtains a certain score on each objective while not being dominated by other model instances (not scoring worse on one criterion while not performing better on any other). To derive such model instances, we use NSGA-II (cf.~\cite{deb2002fast}). We run the algorithm $r$ times per patient.

Using these model instances, we compute a score for the candidate model $M$ based on four criteria (see~\cite{van2017assessment} for a more precise formulation and the rationale for each aspect):

%
%
%
%

\subsubsection{Descriptive Capability}

As said, multiple model instances result from the optimization process, each having their own niche (and error score) in terms of the objectives, none being dominated by another. To compute the score for the descriptive capability (i.e. the error on the training set) we compute the hypervolume from the worst point in the error space (an error of 1 on all evaluation criteria, as values are assumed to be normalized) to the Pareto front. This is called the dominated hypervolume. A value of 1 is the best value (i.e. the Pareto front including model instances with all zero error values). We compute the average dominated hypervolume per patient (remember, we run the algorithm $r$ times) and compute the average over all patients $p$ ($\mu_d$) as well as the standard deviation $\sigma_d$. The descriptive score of an individual $M$ (i.e. model) is:

\begin{align}
descriptive\_score_M = \mu_{d}(1-\sigma_{d})
\end{align}

Hence, the higher and more consistent the score is over all patients, the higher the score in terms of descriptive capabilities.

\subsubsection{Predictive Capability}

In order to compute the predictive capability, we apply the model instances we have found for the training data to unseen validation data (and again, predict $n$ steps ahead for as long as we have validation data, starting from the real initial values). This results in an error associated with each criterion for each model instance for each patient. We compute the mean $\mu_{ap}$ and standard deviation $\sigma_{ap}$ over all of these errors and compute the predictive score:

\begin{align}
predictive\_score_M = (1-\mu_{ap})(1-\sigma_{ap})
\end{align}

\subsubsection{Parameter Sensitivity}

Parameters should be useful and have an impact on the performance of the model. A correlation analysis is performed between the parameter values and the error on the training set for each objective. A parameter is deemed useful in case the highest found correlation (of all correlations over all model instances, patients, and objectives) is above 0.35. The total number of useful parameters is then divided by the total number of parameters:

\begin{align}
sensitivity\_score_M = \frac{\sum \limits_{l=1}^{k}{useful_{M,\lambda_l}}}{k}
\end{align}

\subsubsection{Model Complexity}

Finally, model complexity is weighed, the more complex the model, the lower the score for this aspect will be. It is defined by the number of parameters in the model, divided by the maximum (since we always have all states in the model):

\begin{align}
complexity\_score_M = \frac{k}{\lambda_{max}}
\end{align}
\\

\noindent These scores are combined using a weighed sum:

\begin{align}
fitness_M =
w_1 \cdot descriptive\_score_M + 
w_2 \cdot predictive\_score_M + \nonumber\\
w_3 \cdot sensitivity\_score_M + 
w_4 \cdot complexity\_score_M
\end{align}

\section{Dataset}

We want to investigate how well our approach performs compared to alternative models. For this comparison, we require a dataset that includes a substantial number of patients for whom measurements of multiple health states have been performed over time. In our case, we have obtained such a dataset from the domain of mental health. Nowadays, interventions in mental health are becoming more and more digitized. For example, apps are being developed that can aid depressed patients to battle their depression (see e.g. \cite{riper2010theme}). Next to cost effectiveness, such apps also bring benefits when it comes to tracking the health state of patients as people carry their phone with them all the time. The tracking involves asking patients to score various aspects of their mental health on a regular basis (e.g. their mood) using pop ups on their mobile device. This is commonly referred to as Ecological Momentary Assessment (EMA). Our dataset originates from the E-COMPARED project. This project is focused on studying the effectiveness of interventions for depression. Within the project, a comparison is made between treatment as usual and blended care. The blended care setup features an app the depressed patients can use. Table~\ref{tab:ema} shown an overview of the questions posed to the patients. Note that not all questions are posed on a daily basis.

\begin{table}[!htb]
\caption{The EMA questions that are present in the dataset.}
\label{tab:ema}
\begin{center}
\begin{tabular}{p{3cm} | p{8cm}}
\hline
\textbf{Abbreviation} & \textbf{EMA question} \\
\hline
Mood & How is your mood right now? \\
Worry & How much do you worry about things at the moment? \\
Self-Esteem & How good do you feel about yourself right now? \\
Sleep & How did you sleep tonight? \\
Activities done & To what extent have you carried out enjoyable activities today? \\
Enjoyed activities &  How much have you enjoyed the days activities? \\ 
Social contact &  How much have you been involved in social interactions today? \\
\hline
\end{tabular}
\end{center}
\end{table}

We obtain a dataset of 60 patients from the project. These are patients that have a long enough history to make up an interesting time series (at least 40 days of measurements). Some of the questions are posed multiple times a day (the mood), while others are only posed once a day or even less. To create a suitable dataset we: (1) normalize them on a scale in the range $[0,1]$; (2) aggregate the values per question on a daily basis by averaging in the event of multiple answers per day, and (3) in case a question does not have any answer on a day we linearly interpolate it based on the last known value and the first value in the future. 

\begin{figure}[!htb]
\centering
\includegraphics[width=0.6\textwidth]{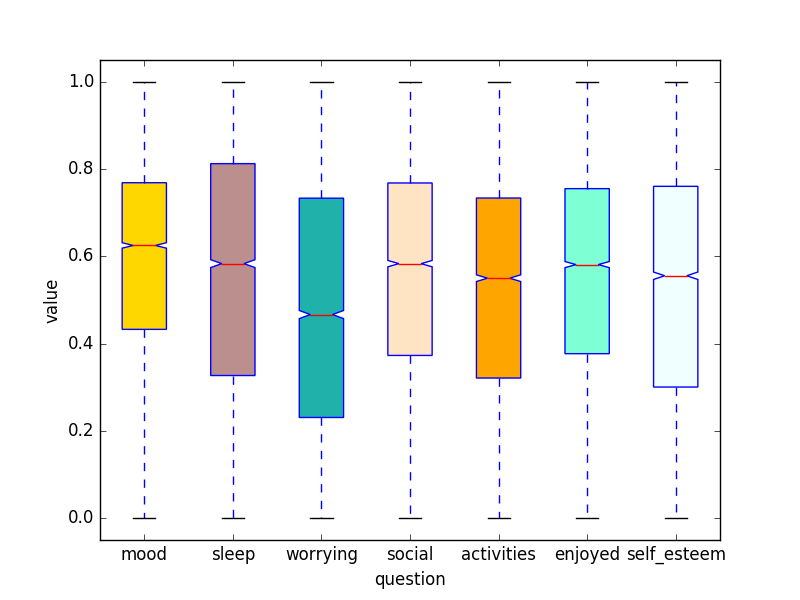}
\caption{Distributions of the responses to the EMA questions.}
\label{fig:datadistr} 
\end{figure}

The resulting dataset contains on average 119.66 days of data per patient, with a large standard deviation of 70.55 days. Figure~\ref{fig:datadistr} shows a boxplot covering the different questions and the distribution of the answers the patients gave. Mood shows the narrowest distribution while the sleep and worrying questions seem to have the largest spread in answers.

\section{Experimental Setup}

In our experimental setup, two phases are distinguished: (1) exploring hyperparameters of the GP framework and generating a dynamical systems model with the best settings, and (2) comparing the performance of the resulting model with alternative approaches. Before diving into those details, we first explain how we further prepare the dataset.

\subsection{Dataset preparation}

We select all data from a random sample of 30 patients as an \emph{in sample} dataset and the 30 remaining patients as \emph{out of sample} dataset. The data of each patient is split in a training, validation, and test set, containing 60, 20, and 20\% of the data respectively, in a time ordered fashion. We evaluate the performance for two states in the dataset: \emph{mood} and \emph{sleep}, both deemed highly relevant by domain experts. The others are less important to predict in the future, but could contain important predictive information. The root mean squared is used as an evaluation metric. We predict 1, 2, and 3 time points (i.e. days) ahead. 

\subsection{Parameter Settings and GP runs}

GP-HD contains a number of hyperparameters. An overview is shown in Table~\ref{tab:params}. Some hyperparameters we fix based on the literature (those that have been shown to work well for multiple types of problems), while we study the influence of others. 

\begin{table*}[!htb]
\caption{Parameter settings for GP-HD.}
\label{tab:params}
\begin{center}
\begin{tabular}{p{1.5cm}|p{6.5cm}|p{3.5cm}}
\hline
\textbf{Parameter} & \textbf{Meaning} & \textbf{Values} \\
\hline\multicolumn{3}{c}{\emph{Fixed}} \\
\hline $p_{op}$ & Probability to select operator in random generation of (sub)trees & 0.5 (cf.~\cite{cao2000evolutionary}) \\
\hline $t_{s}$ & Tournament size & 3 (cf. \cite{cao2000evolutionary})\\
\hline $r_{max}$ & number of NSGA-II runs per patient & 3 (cf.~\cite{van2017assessment}) \\
\hline $p_{r}$ & probability of retaining a copy of a parent & 0.1 (cf.~\cite{cao2000evolutionary}) \\
\hline $\phi$ & Number of tries for crossover & 3 (cf.~\cite{cao2000evolutionary})\\
\hline $w_1,\dots,w_4$ & Weight of evaluation criteria & All set to 0.25 (cf.~\cite{van2017assessment}) \\
\hline\multicolumn{3}{c}{\emph{Varied}} \\
\hline $\lambda_{max}$ & Maximum number of parameters & various (7 selected) \\
\hline $d_{max}$ & Maximum depth of tree that represents model & various (6 selected) \\
\hline $pop_{nsga\_II}$ & population size of the NSGA-II algorithm & \{5, 10, 20, 50\} \\
\hline $gen_{nsga\_II}$ & number of generations for the NSGA-II algorithm & \{5, 10, 20, 50\} \\
\hline $pop_{gp}$ & population size of the GP & \{50, 100\} \\
\hline $gen_{gp}$ & number of generations of the GP & \{50, 100\} \\
\hline
\end{tabular}
\end{center}
\end{table*}

To optimize the varied hyperparameters of the NSGA-II algorithm we perform a number of runs with a small (random) sample of 10 patients from the \emph{in sample} dataset. We create a fixed population of individuals and consider the mean and standard deviation of the fitness values we obtain over multiple runs of the NSGA-II algorithm. For the hyperparameters of the GP algorithm, we have based most of the settings on literature and performed some initial runs to set the more problem dependent hyperparameters appropriately. Initial experiments using the \emph{in sample} data with $\lambda_{max}$ and $d_{max}$ showed that one parameter per state (i.e. 7 in total) works best, as well as a maximum depth of six (simple, yet sufficiently powerful models). We study the influence of two hyperparameters in more detail, namely the population size and numbers of generations. We report results of experiments for those hyperparameters and focus on the convergence and the overall quality of the solutions in terms of fitness. We select the best hyperparameter setting, run it 10 times using the \emph{in sample} data and select the best individual (i.e. model) we encounter. We continue with this model in the next phase.

\subsection{Performance evaluation}

We compare the best individual we have generated with three alternative approaches: 
\begin{enumerate}
\item a dynamical systems model we obtained from the literature (cf.~\cite{abro2016validation}, referred to as the \emph{literature model}). This does not predict \emph{sleep} but does include nearly all other states we have in our dataset. It includes a total of 25 parameters.
\item an LSTM model per individual patient (\emph{individual LSTM}).
\item a single LSTM model across all patients (\emph{generic LSTM}).
\end{enumerate}

To train the dynamical systems models (\emph{GP model} and the \emph{literature model}), we optimize the parameters for each patient individually, by applying NSGA-II on the training portion of the data of that patient, and select the model instance for the patient that minimizes the sum of the errors on the validation set. We do this for both the \emph{in sample} and \emph{out of sample} data. Note that the GP model itself has of course been generated using only the \emph{in sample} data.

We have chosen an LSTM model as benchmark machine learning model as this has shown to work best on this type of data (see~\cite{mikus2018predicting}). For the LSTM models we use the combination of the training and validation part of the patient data as training set. Six output neurons are used (we have three future time points we want to predict for two states). We have experimented with various parameter settings of the LSTM in some initial runs, which showed that taking a batch size of 7 combined with a single layers of 128 hidden neurons work best. We train for 30 epochs. Here, the \emph{individual LSTM} is trained and applied per patient (both for the \emph{in sample} and \emph{out of sample} patients), while the \emph{generic LSTM} is trained on all patients in the \emph{in sample} data.

We measure the performance of all algorithms on the test set part of the data of each patient, both for the \emph{in sample} and \emph{out of sample} data. We perform only single runs of the algorithms per patient as the number of patients will allows us to tackle the stochasticity.

\section{Results}

In the results, we first focus on the parameter settings of our GP-HD approach, followed by the resulting behavior of the algorithm and the best model we find. We then compare the performance of that model to the benchmark algorithms.

\subsection{Parameter Settings and GP runs}

\begin{figure}[!htb]
\centering
\includegraphics[width=0.6\textwidth]{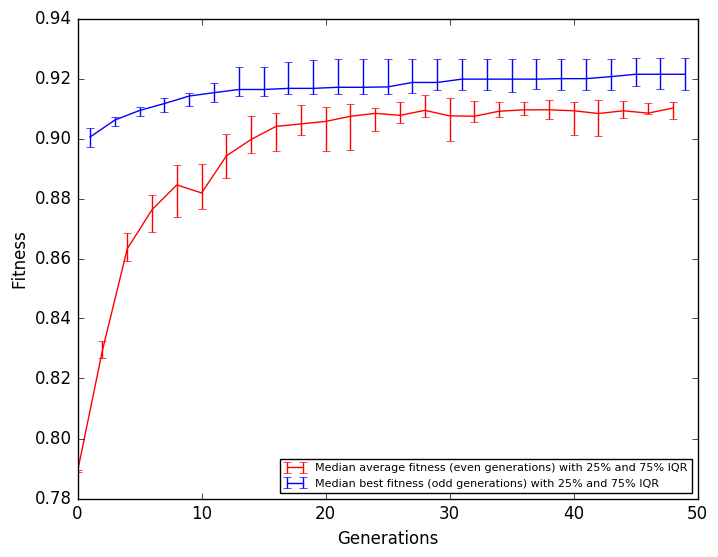}
\caption{Development of the fitness values over the generations over 10 runs, note that the y-axis does not start at 0.}
\label{fig:popmood} 
\end{figure}

When we consider the influence of the hyperparameter settings of NSGA-II on the fitness values found, we hardly see any difference between the different parameter settings we identified, not in terms of the absolute fitness value, nor in terms of the standard deviation. Hence, we select the cheapest option in terms of computation ($pop_{NSGA\_II}=5$ and $gen_{NSGA\_II}=5$). When we look at the hyperparameters of the GP, we see that the process converges before $50$ generations independent of the population size. During the initial runs, the larger population size seemed slightly better, which we therefore use (so, $pop_{gp}=100$ and $gen_{gp}=50$). Figure~\ref{fig:popmood} shows the median fitness values over 10 runs and the 25-and 75\% inter quartile ranges. We see that the fitness nicely converges and obtain best fitness values of around 0.92. 

The best model we obtain has a fitness value of 0.9372 (not remarkably higher compared to other models we observe in the same run or other runs of our framework). The model has the following specification:

\begin{flalign}
&s_{mood}(t+1) = s_{mood}(t)+\gamma_1 \cdot (s_{sleep}(t) \cdot (\gamma_1 - s_{mood}(t)))\\
&s_{sleep}(t+1) = s_{sleep}(t)\\
&s_{worrying}(t+1) = (s_{enjoyed}(t)-s_{social}(t))\cdot \nonumber \\
&\qquad (s_{enjoyed}(t)-s_{self\_esteem}(t))\cdot((s_{enjoyed}(t)- \nonumber \\
&\qquad s_{social}(t))\cdot s_{social}(t))\\
&s_{social}(t+1) = ((s_{worrying}(t)-(\gamma_1-s_{social}(t)))\cdot (s_{worrying}(t)\cdot  \nonumber \\ 
&\qquad s_{sleep}(t)))+(((s_{worrying}(t)-(\gamma_1-s_{social}(t)))\cdot \nonumber \\
&\qquad (s_{sleep}(t)*s_{sleep}(t)))+\gamma_1)\\
&s_{activities}(t+1) = s_{social}(t)\\
&s_{enjoyed}(t+1) = s_{self\_esteem}(t)\\
&s_{self\_esteem}(t+1) = s_{mood}(t)\\\nonumber
\end{flalign}

The model only has one parameter ($\gamma_1$). Furthermore, we see that for \emph{sleep} an extremely simple relationship is found (just take the previous value). While one could argue this does not show the benefit of our approach for this aspect, it does show that it does not generate unnecessarily complex models. For the \emph{mood}, a more complex relationship is observed, drawing advantage of the previous values measured for \emph{sleep}. A relationship between \emph{mood} and \emph{sleep} has been reported in the psychological literature (see e.g.~\cite{thomsen2003rumination}). When comparing the outcome to the literature model (\cite{abro2016validation} only focusing on mood), we see the literature model being much more complex and less insightful (with 25 parameters) while the prediction of mood depends on three factors that are not included in the resulting GP model: the \emph{social} interactions, number of \emph{activities}, and how much the patient \emph{enjoyed} the activities, whereas the literature model does not use \emph{sleep}. A remarkable difference that will be analyzed in more detail with clinical psychologists.

\begin{figure}
\centering
\includegraphics[width=0.6\textwidth]{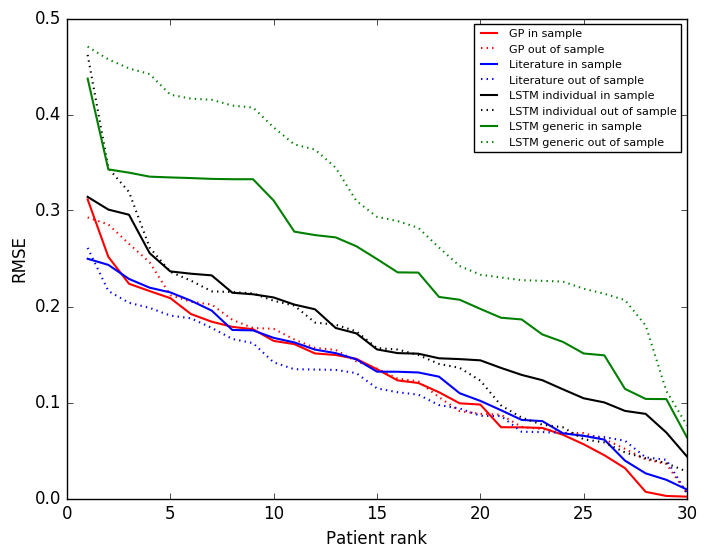}\\
\caption{Ranked RMSE's for prediction of mood at time $t+1$ for all algorithms (in and out of sample)}
\label{fig:errormood} 
\end{figure}

\subsection{Performance Evaluation}

Let us explore how well the model performs compared to other approaches. Table~\ref{tab:rmses} shows the errors we obtain over the different algorithms, evaluation criteria and number of time steps ahead. We have statistically compared the difference in performance of the other algorithms compared to our GP-HD model using a Wilcoxon ranked sum test ($p=0.05$), the results of the significance test are also shown in the table. 

\begin{table*}[!htb]
\caption{Median RMSE performances of algorithms. Scores in \emph{italics} indicates that the performance is significantly worse ($p<0.05$, Wilcoxon ranked sum test) than the performance of the GP, significantly better compared to the GP is not observed.}
\label{tab:rmses}
\begin{center}
\begin{tabular}{|p{1.5cm}|p{1cm}|p{0.8cm}|p{0.8cm}|p{0.8cm}|p{0.8cm}|}
\hline
\textbf{Algorithm} & \textbf{$t +$} & \multicolumn{2}{c|}{\textbf{RMSE (in sample)}} &  \multicolumn{2}{c|}{\textbf{RMSE (out of sample)}}\\\cline{3-6}
& & mood & sleep & mood & sleep \\
\hline
\multirow{3}{*}{GP}   & 1 &  0.129  &  0.030  &  0.130  &  0.078 \\\cline{2-6}
 & 2 &  0.162  &  0.068  &  0.148  &  0.122 \\\cline{2-6}
 & 3 &  0.173  &  0.093  &  0.169  &  0.158 \\\hline
\multirow{3}{*}{Literature}   & 1 &  0.132  &  -  &  0.113  &  - \\\cline{2-6}
 & 2 &  0.149  &  -  &  0.148  &  - \\\cline{2-6}
 & 3 &  0.174  &  -  &  0.172  &  - \\\hline
\multirow{3}{*}{LSTM ind.}   & 1 &  \emph{0.153}  &  \emph{0.189}  &  0.156  &  \emph{0.198} \\\cline{2-6}
 & 2 &  0.155  &  \emph{0.190}  &  0.167  &  \emph{0.207} \\\cline{2-6}
 & 3 &  0.154  &  \emph{0.185}  &  0.184  &  \emph{0.226} \\\hline
\multirow{3}{*}{LSTM gen.}   & 1 &  \emph{0.242}  &  \emph{0.283}  &  \emph{0.291}  &  \emph{0.302} \\\cline{2-6}
 & 2 &  \emph{0.239}  &  \emph{0.295}  &  \emph{0.265}  &  \emph{0.283} \\\cline{2-6}
 & 3 &  \emph{0.292}  &  \emph{0.254}  &  \emph{0.309}  &  \emph{0.307} \\
\hline
\end{tabular}
\end{center}
\end{table*}

We observe that our approach is not significantly outperformed by any other approach, while it significantly outperforms the generic LSTM in all cases and the individual LSTM for the \emph{sleep} state. The literature model and our GP-HD model perform equally well. Our approach also seems to generalize well (considering the performance on \emph{out of sample} patients), especially for the \emph{mood} state. Figure~\ref{fig:errormood} shows the difference in performance over different patients for \emph{mood} at $t+1$ (other prediction intervals show similar patterns) for both the \emph{in sample} and \emph{out of sample} patients. We can observe similar patterns across all algorithms, though performance of the generic LSTM model is a lot poorer for unseen patients.

%

\begin{figure*}[t!]
\centering
$\begin{array}{rl}
\includegraphics[width=0.45\textwidth]{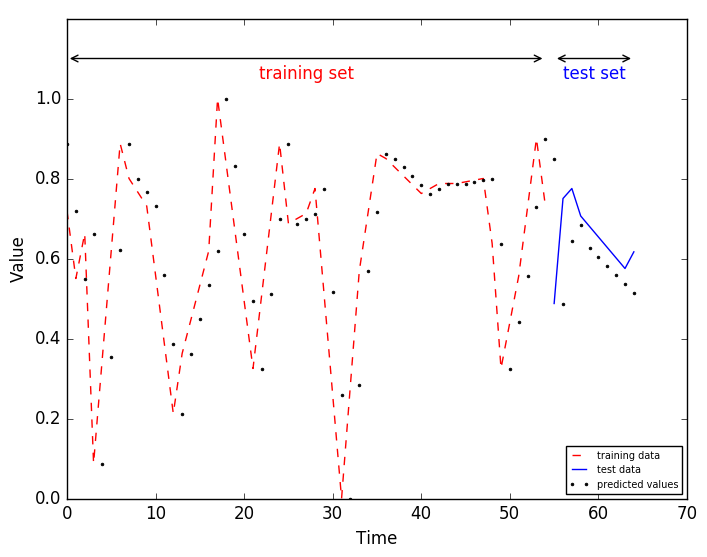} &
\includegraphics[width=0.45\textwidth]{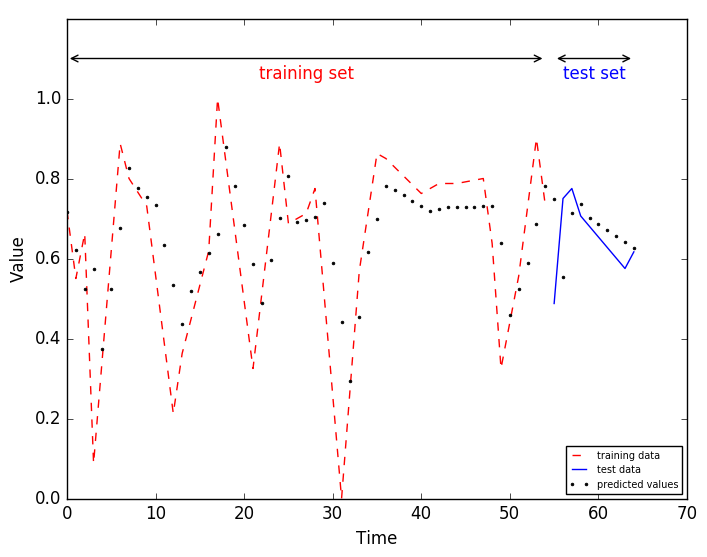} \\
\includegraphics[width=0.45\textwidth]{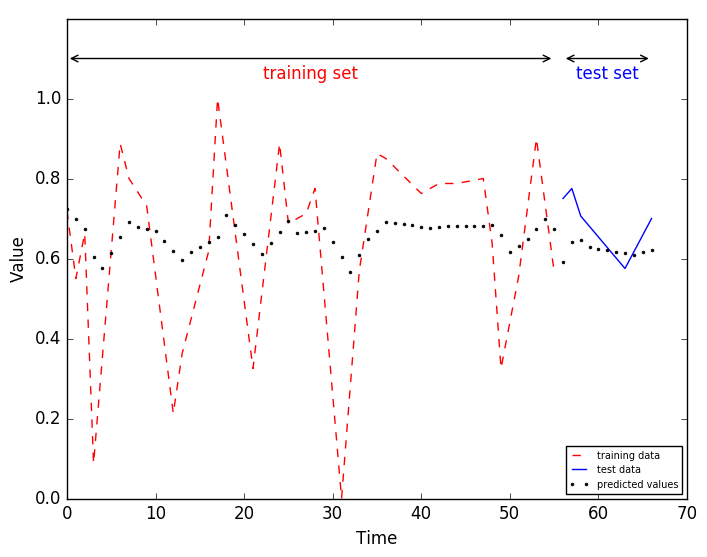} & 
\includegraphics[width=0.45\textwidth]{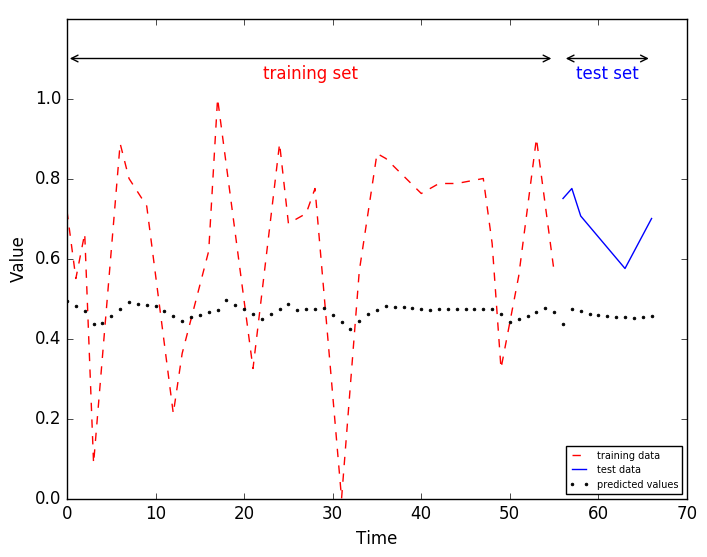} \\
\end{array}$
\caption{Illustration of predictions for mood at time $t+1$  for the out of sample patient with the highest variation in performance (black dots are the predictions of the models). Note that the training set is the combination of the training and validation part of the data. Upper left: GP; upper right: literature, lower left: individual LSTM, lower right: generic LSTM.}
\label{fig:predictions}
\end{figure*}

Finally, Figure~\ref{fig:predictions} shows an example \emph{out of sample} patient (the patient with the highest variation in performance scores) and the accompanying predictions for \emph{mood} at time $t+1$. It can clearly be observed that the GP and literature model predict quite reasonable, while the LSTM individual model does follow the trends, but provides a prediction closer to the average value. This holds even more extreme for the generic LSTM example.

%
%
%

\section{Discussion}

In this paper, we have presented GP-HD to generate dynamical systems models for predicting developments of health states over time. This approach tailored an existing GP approach (cf.~\cite{cao2000evolutionary}) using a fitness function based on an evaluation framework for dynamical systems models (cf.~\cite{van2017assessment}). The research question we posed was: Is the predictive performance of models generated by GP-HD better than state-of-the-art data-driven and knowledge-driven approaches for a real-life case study in the health domain? Based on the results we have obtained for the mental health case studied in this paper we can answer this question with a partial yes. The approach is on par with a literature based dynamical systems model (while being a simpler and more insightful model), outperforms a generic LSTM model, and scores at least as well as individual LSTM models. Of course, it is hard to generalize these results over other datasets. The relationships that are used in the resulting GP model are in line with literature in the psychological domain (e.g. \cite{thomsen2003rumination}). 

A lot of research has been devoted to data-driven predictive models for health. Specially, a variety of LSTM based approaches have been proposed for modeling temporal data in the health domain (see e.g. \cite{lipton2015learning}). In addition, ways to engineer temporal features are seen (e.g.~\cite{batal2013temporal,hoogendoorn2017machine})). However, none make the combination we present in this paper: a generic model with parameters that can be tailored towards individuals. Of course, more GP-based approaches have been developed (e.g.\cite{schmidt2009distilling}), but none are focused on the specific setting with multiple datasets (one per patient) we have. For the domain of mental health there are only few models that have been developed to provide more fine grained (e.g. daily) predictions of the mental health state, see e.g. \cite{daugherty2009mathematical,likamwa2013moodscope,osmani2013monitoring}. Due to the differences in the characteristics of these groups as well as the measurements performed performances are difficult to compare. 

For future work, we want to apply the proposed approach to other health datasets and explore the influence of the hyperparameter settings of the evaluation approach more rigorously. We also want to make the framework more efficient using racing (cf.~\cite{maron1997racing}). In addition, we want to study how we can improve the LSTM performance further. The \emph{generic LSTM} only has access to the \emph{in sample} data while the  \emph{individual LSTM} can only use the data of the specific patient. Studying a hybrid solution of the LSTM (e.g. using transfer learning) where we tailor the generic LSTM model based on a small portion of the data of a specific patient is therefore a next step we want to take. 

\section*{Acknowledgements}

This research has been performed in the context of the EU FP7 project E-COMPARED (project number 603098). We want to thank all trial coordinators in the countries where the MoodBuster platform has been used.

\bibliography{mybib}
\bibliographystyle{splncs04}
\end{document}